\documentclass{article}

    \PassOptionsToPackage{numbers, compress}{natbib}
\usepackage{subcaption}
\usepackage[dblblindworkshop, final]{neurips_2026}
\workshoptitle{Data Attribution and Provenance (ATTRIB)}
\usepackage[utf8]{inputenc} 
\usepackage[T1]{fontenc}    
\usepackage{hyperref}       
\usepackage{url}            
\usepackage{booktabs}       
\usepackage{amsfonts}       
\usepackage{nicefrac}       
\usepackage{microtype}      
\usepackage{xcolor}         
\usepackage{graphicx}
\graphicspath{{figures/}}

\title{Form Over Content In Gradient-Based Data Attribution Methods}

\author{
  Sunwoo Kim \quad Seokwon Jung \quad Sohyung Kim \quad Seong Joon Oh \quad Alice Oh \\
  KAIST \\
  \texttt{sunwoo.kim@kaist.ac.kr} \quad
  \texttt{tjrdnjs0313@kaist.ac.kr} \quad
  \texttt{ks000225@gmail.com} \\
  \texttt{coallaoh@kaist.ac.kr} \quad
  \texttt{alice.oh@kaist.edu}
}

\begin{document}

\maketitle

\begin{abstract}

Data attribution methods using gradient similarity are widely used to analyze and select training data for large language models, but what gradient similarity actually measures is debated.
Some interpret it as identifying task-relevant skills, while other work report that surface form is the main factor.
We resolve this debate for supervised fine-tuning examples by varying task and answer format independently.
Specifically, we render benchmarks in different answer formats, such that datasets can share a task without a format or a format without a task.
We find that gradient alignment follows the answer format, as benchmark pairs sharing an answer format align strongly (disattenuated cosine near 0.4), while same benchmarks rendered with different answer format classes show no alignment (near 0.0).
We demonstrate that this ordering holds from the earliest pretraining checkpoints through post-training, and across model scales and families.
We then analyze the released selections of LESS, a gradient-based data selection method for instruction tuning, and find that each target's selections over-represent the target's own answer format.
Hence, we demonstrate that gradient-based attribution methods track format similarity more than task semantics, meaning that such methods, as well as the semantic interpretation of the gradient, should be tested on data where answer format and task vary independently for greater robustness and reliability.
\end{abstract}

\section{Introduction}
Various data attribution methods have been developed to improve the analysis of training data for deep learning models such as transformer-based large language models (LLMs). 
Many such methods leverage the loss gradient to measure the influence of training data \citep{grosse2023studyinglargelanguagemodel, choe2025what, park2023trak, xia2024less, NEURIPS2020_e6385d39}.
A representative method upon which many other methods are based on is TracIn, which uses the raw similarity, or dot product, between loss gradients of data points to calculate the influence of a training data point on a test data point \citep{NEURIPS2020_e6385d39}.

The literature disagrees on why these methods work and what the gradient actually represents.
\citet{xia2024less} state that the gradient signal ``goes beyond surface form cues to identify data that exemplifies the necessary reasoning skills,'' while \citet{wang2024twostage} report that answer format dominates over the semantic content of the task in the fine-tuning gradient.
The confusion has persisted because format and semantics are spuriously correlated in data, especially benchmarks, so observation alone cannot separate them;
for example, factual knowledge benchmarks tend to share the same multiple-choice format, so a gradient that matches format and a gradient that matches knowledge make the same predictions.
The confusion is pertinent to \textit{targeted instruction tuning}, where the gradient signal is used to pick training examples intended to exemplify the capabilities a target requires, based on the assumption that the gradient encodes semantics \citep{xia2024less}.
We resolve the confusion by varying the \textit{task}\footnote{Note that we use the terms \textit{task} and \textit{benchmark} interchangeably, as the benchmarks used assess unique capabilities.} and the \textit{answer format} of the same questions independently, and measuring each factor's effect on gradient alignment.

Our contributions are threefold:
\begin{itemize}
  \item We show that answer format, not task content, governs gradient similarity in LLMs, meaning that influence as measured by gradient-based data attribution methods does not imply task relatedness.
  \item We find that such dominance of format over content is consistent across pre-training and post-training steps, as well as across model scales. 
  \item We analyze the selection resulting from an existing gradient-based data selection method, namely LESS \citep{xia2024less}, to demonstrate that the resulting selection is skewed towards same-format samples. 
\end{itemize}

\section{Related Work}

\paragraph{Gradient-based attribution and its semantic reading.}
The representative gradient-based data attribution method is TracIn, which estimates the influence of a training example on a test example by accumulating dot products of their loss gradients over training checkpoints \citep{NEURIPS2020_e6385d39}. 
Later work scales this approach to larger models and corpora \citep{park2023trak, grosse2023studyinglargelanguagemodel, choe2025what, chang2025scalable}.
LESS applies it to data selection and interprets the signal as identifying data that exemplifies the necessary reasoning skills \citep{xia2024less}.
This semantic reading of gradient geometry is widely shared:
Gradients are described as task-sensitive representations of knowledge \citep{zhao-etal-2025-beyond-similarity}, and gradient clusters are treated as task experts \citep{li2025ensembles}, task-agnostic coresets \citep{zhang-etal-2025-tagcos}, separable basic abilities \citep{wang2026decomposing}, or attributions of reasoning ability \citep{kou2025which}.
The closest supporting evidence, which are influence analyses that find increasingly abstract relationships between influential documents and model generations \citep{ruis2025procedural, grosse2023studyinglargelanguagemodel}, concerns a different object: The influence of pretraining documents on model generations, not the gradients of supervised fine-tuning examples.
In short, there is prior work that assumes that gradient alignment between fine-tuning examples reflects shared task content, and we test this assumption.

\paragraph{Evidence for surface form.}
Other research suggests the gradient encodes surface format more strongly than content.
At the outcome level, LESS fails to beat random selection at large pool scales \citep{xia-etal-2025-rethinking}, lexical baselines remain competitive with gradient attribution at fact tracing \citep{chang2025scalable}, and gradient similarity fails to indicate task relatedness in multi-task classification \citep{ni-etal-2023-aggregating}.
At the signal level, the only direct gradient measurement is confined to a single dataset: 
\citet{wang2024twostage} shuffle one dataset's labels and find that early fine-tuning gradients track the format.
Whether format organizes the gradients \textit{between} datasets, the comparison that attribution and selection actually perform, remains unmeasured.
The confound surfaces in deployed pipelines as well.
Influence magnitudes track completion length inside LESS itself \citep{dai-etal-2025-improving}, gradient-matched selections of harmful data turn out to be lists, bullet points, and math questions \citep{he2024what}, and influence retrieval for verbalized confidence latches onto surface markers of certainty rather than content \citep{10.1007/978-3-032-21289-4_34}.
We isolate the confound with a controlled design, varying task and answer format independently in the setting attribution methods operate in, where candidate and target come from different datasets.

\begin{figure}[t]
  \centering
  \includegraphics[width=\linewidth]{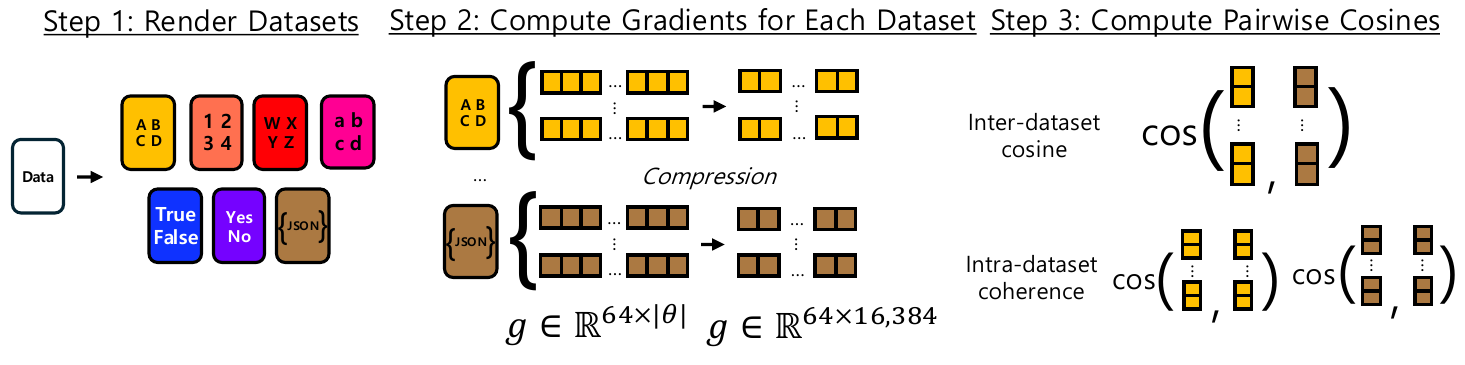}
  \caption{Overview of method used to extract pairwise cosine similarities between and within datasets.}
  \label{fig:method_figure_cropped}
\end{figure}

\section{Method} 
\subsection{Preliminaries}
\label{preliminaries}
\paragraph{Per-example loss gradients.}
Let $f_\theta$ be a language model with parameters $\theta$, and let a supervised example $z=(x,y)$ consist of a prompt $x$ and an answer $y$.
The loss $\ell(z;\theta)$ is the mean cross-entropy of the answer tokens given the prompt, and the per-example gradient is $g(z;\theta) = \nabla_\theta\, \ell(z;\theta)$.
Because the loss is masked to the answer span, $g$ is the update that the example would contribute during instruction tuning, and it is the basic object of gradient-based attribution.

\paragraph{Gradient-based influence.}
TracIn \citep{NEURIPS2020_e6385d39} estimates the influence of a training example on a test example as a learning-rate-weighted sum of gradient dot products over training checkpoints $\theta_1,\dots,\theta_K$.
LESS \citep{xia2024less} adapts this score to instruction tuning:
Notably, gradient alignments are calculated by cosine so that long answers do not dominate through gradient norm. 
The training pool is ranked by the aggregated similarity score with a target set across checkpoints, and the top 5\% is selected for fine-tuning.
We analyze the selections this method produces in Section~\ref{less_posthoc}.

\subsection{Our Method: Measuring Dataset-wise Gradient Alignment}

We measure gradient alignment between datasets in three steps as shown in Figure~\ref{fig:method_figure_cropped}.

\paragraph{Step 1: Render datasets.}
We create a \textit{dataset} by rendering the questions of a benchmark into one answer format, keeping the questions themselves fixed.
We cross five benchmarks, TriviaQA \citep{joshi-etal-2017-triviaqa}, SQuAD \citep{rajpurkar-etal-2016-squad}, GSM8K \citep{cobbe2021gsm8k}, ARC-Challenge \citep{clark2018thinksolvedquestionanswering}, and CRUXEval \citep{pmlr-v235-gu24c}, with seven answer formats spanning three structure classes: binary judgment (True/False, Yes/No), four-option selection (A--D, 1--4, W--Z, a--d), and structured generation (JSON); details in Appendix~\ref{app:tables}.
The resulting 35 datasets let any two be compared while changing only the task or only the answer format.

\paragraph{Step 2: Compute gradients for each dataset.}
For $n = 64$ examples per dataset we compute the per-example answer-span gradient $g(z;\theta)$ of Section~\ref{preliminaries}, taken with respect to the transformer body only; excluding the embedding and unembedding tables ensures that vocabulary overlap between answer tokens cannot drive similarity.
Each gradient is compressed with a seeded CountSketch projection \citep{10.1145/1553374.1553516} to $k = 16{,}384$ dimensions, which preserves inner products in expectation.
Each dataset becomes a matrix in $\mathbb{R}^{64 \times 16{,}384}$.

\paragraph{Step 3: Compute pairwise cosines.}
The \textit{inter-dataset} alignment of datasets $A$ and $B$ is the mean cosine over all cross-dataset example pairs:
\begin{equation}
\hat{c}(A,B) \;=\; \frac{1}{|A|\,|B|}\sum_{z \in A}\ \sum_{z' \in B} \cos\!\big(g(z;\theta),\, g(z';\theta)\big).
\label{eq:pairmean}
\end{equation}
The same statistic measured within a dataset gives the \textit{intra-dataset} coherence $c_A = \hat{c}(A,A)$, evaluated over distinct question pairs.
We disattenuate by coherence, as datasets differ in how noisy their example gradients are, so raw values of Equation~\ref{eq:pairmean} are not comparable across pairs.
If each example gradient is a dataset-level signal $\mu$ plus independent noise, the measured mean is attenuated by exactly $\sqrt{c_A c_B}$, so dividing it out recovers the alignment of the signals themselves \citep{ca468a70-0be4-389a-b0b9-5dd1ff52b33f}:
\begin{equation}
\mathbb{E}\,\hat{c}(A,B)
  \;=\; \cos(\mu_A,\mu_B)\,\sqrt{c_A c_B}
\quad\Longrightarrow\quad
\widetilde{c}(A,B) \;=\; \frac{\hat{c}(A,B)}{\sqrt{c_A c_B}}.
\label{eq:disatt}
\end{equation}
More details on disattenuation may be found in Appendix~\ref{app:disatt}.
As a reference for zero, the 280 dataset pairs that share neither task nor format class have mean $-0.001$ and standard deviation $0.008$ (maximum $0.03$), so we read values within $\pm 0.02$ as null. 
Note, error bars in figures are $\pm 1$ standard error across benchmark identities, so that they reflect whether the effect is consistent across tasks rather than estimator noise (Appendix~\ref{app:errorbars}).

\subsection{Experimental Setup}

\paragraph{Models.}
To test whether the format ordering is a property of fully trained models only, we repeat the measurement at ten OLMo pretraining checkpoints (1B to 4T tokens) \citep{olmo20252olmo2furious}, at ten Pythia checkpoints \citep{pmlr-v202-biderman23a}, and at the base, SFT, DPO, and Instruct stages of OLMo.
To test whether it is specific to a scale or model family, we repeat it at three OLMo scales (1B, 7B, 13B) and on Llama 3.1 \citep{grattafiori2024llama3herdmodels} and Qwen3 \citep{yang2025qwen3technicalreport} models, both base and instruction-tuned.

\paragraph{LESS selections.}
We analyze the selections released by \citet{xia2024less} at \href{https://huggingface.co/datasets/princeton-nlp/less_data}{princeton-nlp/less\_data}: The top 5\% (13,533 examples) of a 270K-example pool for each of three targets (MMLU \citep{hendrycks2021measuring}, BBH \citep{suzgun-etal-2023-challenging}, TydiQA \citep{10.1162/tacl_a_00317}) and three seeds.
We label every selected example and the pool by answer format with a rule-based classifier (\textasciitilde90\% agreement with manual labels on a held-out sample), and report enrichment as the share of a format among selected examples divided by its share of the pool. 
Refer to Appendix~\ref{app:classifier} for more details on the classifier and Appendix~\ref{app:pool} for more details on pool composition.

\section{Results}
\subsection{Answer Format, Not Task, Governs Gradient Alignment}
\begin{figure}[t]
  \centering
  \begin{subfigure}[b]{0.55\linewidth}
    \centering
    \includegraphics[width=0.9\linewidth]{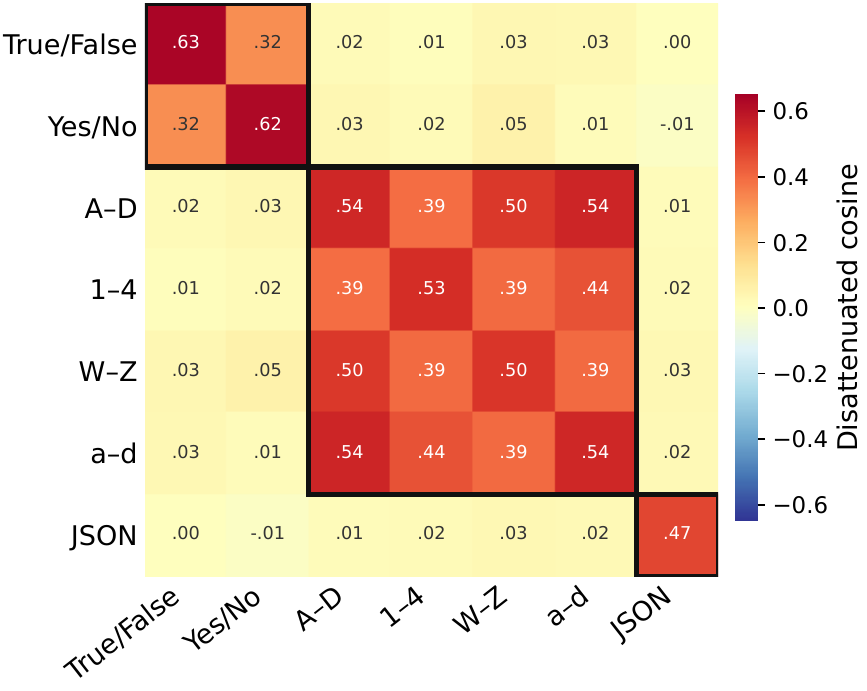}
    \caption{}\label{fig:grid}
  \end{subfigure}\hfill
  \begin{subfigure}[b]{0.43\linewidth}
    \centering
    \includegraphics[width=0.9\linewidth]{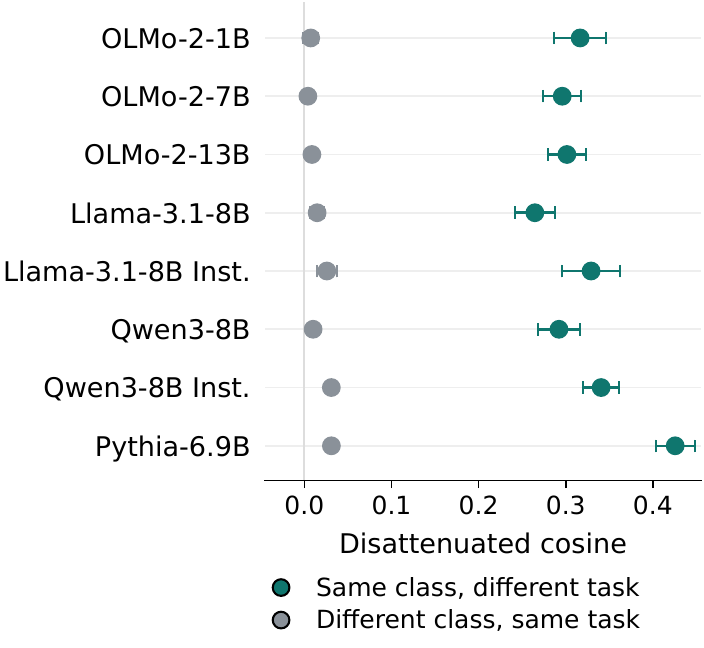}
    \caption{}\label{fig:robustness}
  \end{subfigure}
  \caption{Gradient alignment follows answer format, not task.
  (a) Disattenuated cosine between the seven answer formats, averaged over benchmarks and over eight models from four families. Diagonal cells compare the same format across different benchmarks; off-diagonal cells compare different formats of the same benchmark. Black boxes mark the three answer-structure classes. Alignment is high within a class regardless of task, and near zero across classes even for the same task.
  (b) The same two contrasts measured per model, spanning 1B to 13B and four families.}
  \label{fig:main}
\end{figure}

Figure~\ref{fig:grid} shows cross-dataset cosine similarity while Figure~\ref{fig:robustness} shows the same, aggregated for each model. 
The heatmap shows that same format class dataset pairs have high alignment, while dataset pairs with different format classes have near zero alignment. 
This implies that the gradient encodes more for answer format than task semantics. 
Figure~\ref{fig:robustness} shows that format dominance in the gradient is consistent across model scale and family. 

\subsection{Format Dominance Holds Throughout Pretraining and Post-Training}
\begin{figure}[t]
  \centering
  \includegraphics[width=0.7\linewidth]{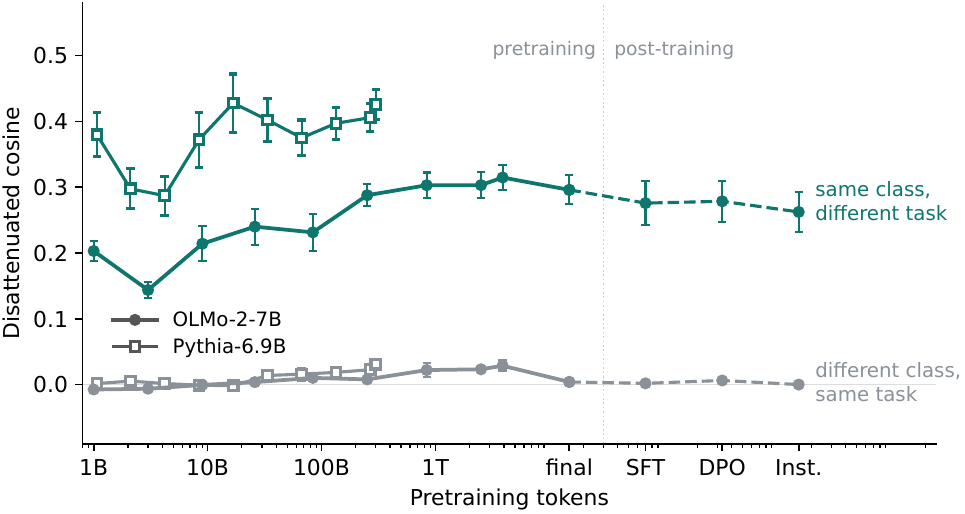}
  \caption{Answer form predicts gradient similarity regardless of the amount of training.}
  \label{fig:fig_trajectory_overlay}
\end{figure}

We show that surface form dominance in gradient similarity is not an artifact of the amount of model pre-training or post-training in Figure~\ref{fig:fig_trajectory_overlay}. 
The figure shows that different benchmark, same answer class cosine similarity stays above the same benchmark, different answer class similarity.

\subsection{LESS Selects for the Target's Answer Format}
\label{less_posthoc}
\begin{figure}[t]
  \centering
  \includegraphics[width=0.6\linewidth]{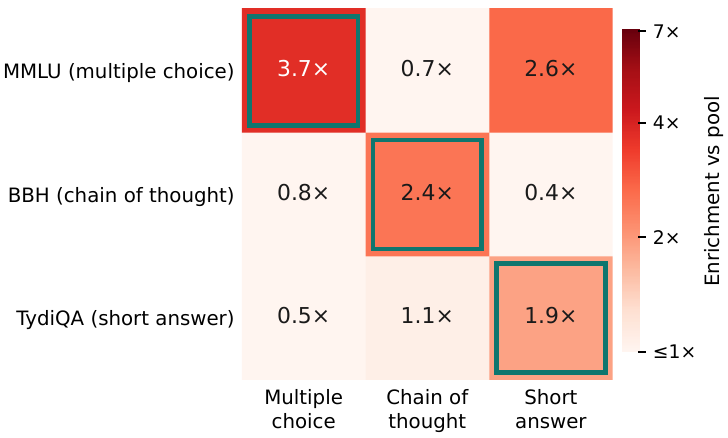}
  \caption{LESS's released selections, by answer format, for the three target formats.
  Each cell is the share of a format among a target's selections divided by its share of the pool; teal boxes mark each target's own format.}
  \label{fig:fig_less_posthoc}
\end{figure}

We analyze the selections that \citet{xia2024less} released for each target benchmark as shown in Figure~\ref{fig:fig_less_posthoc}.
Although the pool contains no data from any of the three target benchmarks, each target's selections are enriched in the target's own answer format: Letter answers are selected at 3.7 times their pool share for MMLU, chain-of-thought at 2.4 times for BBH, and short answers at 1.9 times for TydiQA. 
Figure~\ref{fig:posthoc_full} includes analysis of other answer formats.
The pattern is stable across the three released seeds.

\section{Discussion \& Future Work}
Our results bear on the reliability and robustness of gradient-based attribution and selection:
Their efficacy may partly be an artifact of format coinciding with semantics in current evaluation settings, and may not transfer to pools where the two decouple.
More broadly, researchers should test gradient-based data attribution and selection methods against format-controlled baselines before relying on them.
Future work includes interventional studies: 
For example, whether unifying a pool's answer format makes selection more content-driven and fine-tuning more effective.
Additionally, using learned parameter weightings that separate style from content may remove format dominance.

\section{Limitations}
We assume that task relatedness is a desirable factor in data attribution and especially selection. 
However, format-matched data often genuinely improves benchmark scores in the context of targeted instruction tuning;
we falsify the semantic \textit{interpretation} behind gradient-based selection to question its robustness, not its measured gains.
Additionally, the scope of our experiments may be enlarged: The measurement basis is five English question-answering benchmarks and seven answer formats in three structure classes, with JSON as the only structured-generation member. 
The answer format classifications are also our choices rather than the result of rigorous analysis. 

\begin{ack}
This work was supported by Institute of Information \& communications Technology Planning \& Evaluation(IITP) grant funded by the Korea government(MSIT) (No. RS-2024-00509258 and No. RS-2024-00469482, Global AI Frontier Lab) and by Artificial intelligence industrial convergence cluster development project funded by the Ministry of Science and ICT(MSIT, Korea)\&Gwangju Metropolitan City.
\end{ack}

\bibliography{references}

@inproceedings{xia2024less,
   title={{LESS}: Selecting Influential Data for Targeted Instruction Tuning},
   author={Xia, Mengzhou and Malladi, Sadhika and Gururangan, Suchin and Arora, Sanjeev and Chen, Danqi},
   booktitle={International Conference on Machine Learning (ICML)},
   year={2024}
}

@inproceedings{
wang2024twostage,
title={Two-stage {LLM} Fine-tuning with Less Specialization and More Generalization},
author={Yihan Wang and Si Si and Daliang Li and Michal Lukasik and Felix Yu and Cho-Jui Hsieh and Inderjit S Dhillon and Sanjiv Kumar},
booktitle={The Twelfth International Conference on Learning Representations},
year={2024},
url={https://openreview.net/forum?id=pCEgna6Qco}
}

@inproceedings{
chang2025scalable,
title={Scalable Influence and Fact Tracing for Large Language Model Pretraining},
author={Tyler A. Chang and Dheeraj Rajagopal and Tolga Bolukbasi and Lucas Dixon and Ian Tenney},
booktitle={The Thirteenth International Conference on Learning Representations},
year={2025},
url={https://openreview.net/forum?id=gLa96FlWwn}
}

@misc{grosse2023studyinglargelanguagemodel,
      title={Studying Large Language Model Generalization with Influence Functions}, 
      author={Roger Grosse and Juhan Bae and Cem Anil and Nelson Elhage and Alex Tamkin and Amirhossein Tajdini and Benoit Steiner and Dustin Li and Esin Durmus and Ethan Perez and Evan Hubinger and Kamilė Lukošiūtė and Karina Nguyen and Nicholas Joseph and Sam McCandlish and Jared Kaplan and Samuel R. Bowman},
      year={2023},
      eprint={2308.03296},
      archivePrefix={arXiv},
      primaryClass={cs.LG},
      url={https://arxiv.org/abs/2308.03296}, 
}

@inproceedings{xia-etal-2025-rethinking,
    title = "Rethinking Data Selection at Scale: Random Selection is Almost All You Need",
    author = "Xia, Tingyu  and
      Yu, Bowen  and
      Dang, Kai  and
      Yang, An  and
      Wu, Yuan  and
      Tian, Yuan  and
      Chang, Yi  and
      Lin, Junyang",
    editor = "Christodoulopoulos, Christos  and
      Chakraborty, Tanmoy  and
      Rose, Carolyn  and
      Peng, Violet",
    booktitle = "Findings of the Association for Computational Linguistics: EMNLP 2025",
    month = nov,
    year = "2025",
    address = "Suzhou, China",
    publisher = "Association for Computational Linguistics",
    url = "https://aclanthology.org/2025.findings-emnlp.146/",
    doi = "10.18653/v1/2025.findings-emnlp.146",
    pages = "2698--2711",
    ISBN = "979-8-89176-335-7"
}

@inproceedings{
choe2025what,
title={What is Your Data Worth to {GPT}? {LLM}-Scale Data Valuation with Influence Functions},
author={Sang Keun Choe and Hwijeen Ahn and Juhan Bae and Kewen Zhao and Youngseog Chung and Adithya Pratapa and Willie Neiswanger and Emma Strubell and Teruko Mitamura and Jeff Schneider and Eduard Hovy and Roger Baker Grosse and Eric P. Xing},
booktitle={The Thirty-ninth Annual Conference on Neural Information Processing Systems},
year={2025},
url={https://openreview.net/forum?id=zPKeJAEo27}
}

@inproceedings{park2023trak,
  title = {TRAK: Attributing Model Behavior at Scale},
  author = {Sung Min Park and Kristian Georgiev and Andrew Ilyas and Guillaume Leclerc and Aleksander Madry},
  booktitle = {International Conference on Machine Learning (ICML)},
  year = {2023}
}

@inproceedings{NEURIPS2020_e6385d39,
 author = {Pruthi, Garima and Liu, Frederick and Kale, Satyen and Sundararajan, Mukund},
 booktitle = {Advances in Neural Information Processing Systems},
 editor = {H. Larochelle and M. Ranzato and R. Hadsell and M.F. Balcan and H. Lin},
 pages = {19920--19930},
 publisher = {Curran Associates, Inc.},
 title = {Estimating Training Data Influence by Tracing Gradient Descent},
 url = {https://proceedings.neurips.cc/paper_files/paper/2020/file/e6385d39ec9394f2f3a354d9d2b88eec-Paper.pdf},
 volume = {33},
 year = {2020}
}

@inproceedings{zhao-etal-2025-beyond-similarity,
    title = "Beyond Similarity: A Gradient-based Graph Method for Instruction Tuning Data Selection",
    author = "Zhao, Yang  and
      Du, Li  and
      Ding, Xiao  and
      Ouyang, Yangou  and
      Wang, Hepeng  and
      Xiong, Kai  and
      Gao, Jinglong  and
      Sun, Zhouhao  and
      Xu, Dongliang  and
      Yang, Qing  and
      Li, Dongchen  and
      Qin, Bing  and
      Liu, Ting",
    editor = "Che, Wanxiang  and
      Nabende, Joyce  and
      Shutova, Ekaterina  and
      Pilehvar, Mohammad Taher",
    booktitle = "Proceedings of the 63rd Annual Meeting of the Association for Computational Linguistics (Volume 1: Long Papers)",
    month = jul,
    year = "2025",
    address = "Vienna, Austria",
    publisher = "Association for Computational Linguistics",
    url = "https://aclanthology.org/2025.acl-long.1189/",
    doi = "10.18653/v1/2025.acl-long.1189",
    pages = "24391--24404",
    ISBN = "979-8-89176-251-0"
}

@inproceedings{
li2025ensembles,
title={Ensembles of Low-Rank Expert Adapters},
author={Yinghao Li and Vianne R. Gao and Chao Zhang and MohamadAli Torkamani},
booktitle={The Thirteenth International Conference on Learning Representations},
year={2025},
url={https://openreview.net/forum?id=l0gZS0sAlf}
}

@inproceedings{zhang-etal-2025-tagcos,
    title = "{TAGCOS}: Task-agnostic Gradient Clustered Coreset Selection for Instruction Tuning Data",
    author = "Zhang, Jipeng  and
      Qin, Yaxuan  and
      Pi, Renjie  and
      Zhang, Weizhong  and
      Pan, Rui  and
      Zhang, Tong",
    editor = "Chiruzzo, Luis  and
      Ritter, Alan  and
      Wang, Lu",
    booktitle = "Findings of the Association for Computational Linguistics: NAACL 2025",
    month = apr,
    year = "2025",
    address = "Albuquerque, New Mexico",
    publisher = "Association for Computational Linguistics",
    url = "https://aclanthology.org/2025.findings-naacl.264/",
    doi = "10.18653/v1/2025.findings-naacl.264",
    pages = "4686--4701",
    ISBN = "979-8-89176-195-7"
}

@inproceedings{
kou2025which,
title={Which Data Attributes Stimulate Math and Code Reasoning? An Investigation via Influence Functions},
author={Siqi Kou and Qingyuan Tian and Hanwen Xu and Zihao Zeng and Zhijie Deng},
booktitle={The Thirty-ninth Annual Conference on Neural Information Processing Systems},
year={2025},
url={https://openreview.net/forum?id=b7uniOw0sZ}
}

@inproceedings{
wang2026decomposing,
title={Decomposing the Basic Abilities of Large Language Models: Mitigating Cross-Task Interference in Multi-Task Instruct-Tuning},
author={Bing Wang and Ximing Li and Changchun Li and Jinjin Chi and Gang Niu and Masashi Sugiyama},
booktitle={Forty-third International Conference on Machine Learning},
year={2026},
url={https://openreview.net/forum?id=FFAHL32Wok}
}

@inproceedings{dai-etal-2025-improving,
    title = "Improving Influence-based Instruction Tuning Data Selection for Balanced Learning of Diverse Capabilities",
    author = "Dai, Qirun  and
      Zhang, Dylan  and
      Ma, Jiaqi W.  and
      Peng, Hao",
    editor = "Christodoulopoulos, Christos  and
      Chakraborty, Tanmoy  and
      Rose, Carolyn  and
      Peng, Violet",
    booktitle = "Findings of the Association for Computational Linguistics: EMNLP 2025",
    month = nov,
    year = "2025",
    address = "Suzhou, China",
    publisher = "Association for Computational Linguistics",
    url = "https://aclanthology.org/2025.findings-emnlp.373/",
    doi = "10.18653/v1/2025.findings-emnlp.373",
    pages = "7079--7102",
    ISBN = "979-8-89176-335-7"
}

@inproceedings{ni-etal-2023-aggregating,
    title = "When Does Aggregating Multiple Skills with Multi-Task Learning Work? A Case Study in Financial {NLP}",
    author = "Ni, Jingwei  and
      Jin, Zhijing  and
      Wang, Qian  and
      Sachan, Mrinmaya  and
      Leippold, Markus",
    editor = "Rogers, Anna  and
      Boyd-Graber, Jordan  and
      Okazaki, Naoaki",
    booktitle = "Proceedings of the 61st Annual Meeting of the Association for Computational Linguistics (Volume 1: Long Papers)",
    month = jul,
    year = "2023",
    address = "Toronto, Canada",
    publisher = "Association for Computational Linguistics",
    url = "https://aclanthology.org/2023.acl-long.412/",
    doi = "10.18653/v1/2023.acl-long.412",
    pages = "7465--7488"
}

@inproceedings{
he2024what,
title={What is in Your Safe Data? Identifying Benign Data that Breaks Safety},
author={Luxi He and Mengzhou Xia and Peter Henderson},
booktitle={First Conference on Language Modeling},
year={2024},
url={https://openreview.net/forum?id=Hi8jKh4HE9}
}

@InProceedings{10.1007/978-3-032-21289-4_34,
author="Xia, Yuxi
and Schoenegger, Loris
and Roth, Benjamin",
editor="Campos, Ricardo
and Jatowt, Adam
and Lan, Yanyan
and Aliannejadi, Mohammad
and Bauer, Christine
and MacAvaney, Sean
and Anand, Avishek
and Ren, Zhaochun
and Verberne, Suzan
and Bai, Nan
and Mansoury, Masoud",
title="Influential Training Data Retrieval for Explaining Verbalized Confidence of LLMs",
booktitle="Advances in Information Retrieval",
year="2026",
publisher="Springer Nature Switzerland",
address="Cham",
pages="529--547",
isbn="978-3-032-21289-4"
}

@inproceedings{
ruis2025procedural,
title={Procedural Knowledge in Pretraining Drives Reasoning in Large Language Models},
author={Laura Ruis and Maximilian Mozes and Juhan Bae and Siddhartha Rao Kamalakara and Dwaraknath Gnaneshwar and Acyr Locatelli and Robert Kirk and Tim Rockt{\"a}schel and Edward Grefenstette and Max Bartolo},
booktitle={The Thirteenth International Conference on Learning Representations},
year={2025},
url={https://openreview.net/forum?id=1hQKHHUsMx}
}

@article{ca468a70-0be4-389a-b0b9-5dd1ff52b33f,
 ISSN = {00029556},
 URL = {http://www.jstor.org/stable/1412159},
 author = {C. Spearman},
 journal = {The American Journal of Psychology},
 number = {1},
 pages = {72--101},
 publisher = {University of Illinois Press},
 title = {The Proof and Measurement of Association between Two Things},
 urldate = {2026-08-25},
 volume = {15},
 year = {1904}
}

@inproceedings{10.1145/1553374.1553516,
author = {Weinberger, Kilian and Dasgupta, Anirban and Langford, John and Smola, Alex and Attenberg, Josh},
title = {Feature hashing for large scale multitask learning},
year = {2009},
isbn = {9781605585161},
publisher = {Association for Computing Machinery},
address = {New York, NY, USA},
url = {https://doi.org/10.1145/1553374.1553516},
doi = {10.1145/1553374.1553516},
booktitle = {Proceedings of the 26th Annual International Conference on Machine Learning},
pages = {1113–1120},
numpages = {8},
location = {Montreal, Quebec, Canada},
series = {ICML '09}
}

@inproceedings{joshi-etal-2017-triviaqa,
    title = "{T}rivia{QA}: A Large Scale Distantly Supervised Challenge Dataset for Reading Comprehension",
    author = "Joshi, Mandar  and
      Choi, Eunsol  and
      Weld, Daniel  and
      Zettlemoyer, Luke",
    editor = "Barzilay, Regina  and
      Kan, Min-Yen",
    booktitle = "Proceedings of the 55th Annual Meeting of the Association for Computational Linguistics (Volume 1: Long Papers)",
    month = jul,
    year = "2017",
    address = "Vancouver, Canada",
    publisher = "Association for Computational Linguistics",
    url = "https://aclanthology.org/P17-1147/",
    doi = "10.18653/v1/P17-1147",
    pages = "1601--1611"
}

@inproceedings{rajpurkar-etal-2016-squad,
    title = "{SQ}u{AD}: 100,000+ Questions for Machine Comprehension of Text",
    author = "Rajpurkar, Pranav  and
      Zhang, Jian  and
      Lopyrev, Konstantin  and
      Liang, Percy",
    editor = "Su, Jian  and
      Duh, Kevin  and
      Carreras, Xavier",
    booktitle = "Proceedings of the 2016 Conference on Empirical Methods in Natural Language Processing",
    month = nov,
    year = "2016",
    address = "Austin, Texas",
    publisher = "Association for Computational Linguistics",
    url = "https://aclanthology.org/D16-1264/",
    doi = "10.18653/v1/D16-1264",
    pages = "2383--2392"
}

@article{cobbe2021gsm8k,
  title={Training Verifiers to Solve Math Word Problems},
  author={Cobbe, Karl and Kosaraju, Vineet and Bavarian, Mohammad and Chen, Mark and Jun, Heewoo and Kaiser, Lukasz and Plappert, Matthias and Tworek, Jerry and Hilton, Jacob and Nakano, Reiichiro and Hesse, Christopher and Schulman, John},
  journal={arXiv preprint arXiv:2110.14168},
  year={2021}
}

@misc{clark2018thinksolvedquestionanswering,
      title={Think you have Solved Question Answering? Try ARC, the AI2 Reasoning Challenge}, 
      author={Peter Clark and Isaac Cowhey and Oren Etzioni and Tushar Khot and Ashish Sabharwal and Carissa Schoenick and Oyvind Tafjord},
      year={2018},
      eprint={1803.05457},
      archivePrefix={arXiv},
      primaryClass={cs.AI},
      url={https://arxiv.org/abs/1803.05457}, 
}

@InProceedings{pmlr-v235-gu24c,
  title = 	 {{CRUXE}val: A Benchmark for Code Reasoning, Understanding and Execution},
  author =       {Gu, Alex and Roziere, Baptiste and Leather, Hugh James and Solar-Lezama, Armando and Synnaeve, Gabriel and Wang, Sida},
  booktitle = 	 {Proceedings of the 41st International Conference on Machine Learning},
  pages = 	 {16568--16621},
  year = 	 {2024},
  editor = 	 {Salakhutdinov, Ruslan and Kolter, Zico and Heller, Katherine and Weller, Adrian and Oliver, Nuria and Scarlett, Jonathan and Berkenkamp, Felix},
  volume = 	 {235},
  series = 	 {Proceedings of Machine Learning Research},
  month = 	 {21--27 Jul},
  publisher =    {PMLR},
  url = 	 {https://proceedings.mlr.press/v235/gu24c.html}
}

@misc{olmo20252olmo2furious,
      title={2 OLMo 2 Furious}, 
      author={Team OLMo and Pete Walsh and Luca Soldaini and Dirk Groeneveld and Kyle Lo and Shane Arora and Akshita Bhagia and Yuling Gu and Shengyi Huang and Matt Jordan and Nathan Lambert and Dustin Schwenk and Oyvind Tafjord and Taira Anderson and David Atkinson and Faeze Brahman and Christopher Clark and Pradeep Dasigi and Nouha Dziri and Allyson Ettinger and Michal Guerquin and David Heineman and Hamish Ivison and Pang Wei Koh and Jiacheng Liu and Saumya Malik and William Merrill and Lester James V. Miranda and Jacob Morrison and Tyler Murray and Crystal Nam and Jake Poznanski and Valentina Pyatkin and Aman Rangapur and Michael Schmitz and Sam Skjonsberg and David Wadden and Christopher Wilhelm and Michael Wilson and Luke Zettlemoyer and Ali Farhadi and Noah A. Smith and Hannaneh Hajishirzi},
      year={2025},
      eprint={2501.00656},
      archivePrefix={arXiv},
      primaryClass={cs.CL},
      url={https://arxiv.org/abs/2501.00656}, 
}

@InProceedings{pmlr-v202-biderman23a,
  title = 	 {Pythia: A Suite for Analyzing Large Language Models Across Training and Scaling},
  author =       {Biderman, Stella and Schoelkopf, Hailey and Anthony, Quentin Gregory and Bradley, Herbie and O'Brien, Kyle and Hallahan, Eric and Khan, Mohammad Aflah and Purohit, Shivanshu and Prashanth, Usvsn Sai and Raff, Edward and Skowron, Aviya and Sutawika, Lintang and Van Der Wal, Oskar},
  booktitle = 	 {Proceedings of the 40th International Conference on Machine Learning},
  pages = 	 {2397--2430},
  year = 	 {2023},
  editor = 	 {Krause, Andreas and Brunskill, Emma and Cho, Kyunghyun and Engelhardt, Barbara and Sabato, Sivan and Scarlett, Jonathan},
  volume = 	 {202},
  series = 	 {Proceedings of Machine Learning Research},
  month = 	 {23--29 Jul},
  publisher =    {PMLR},
  url = 	 {https://proceedings.mlr.press/v202/biderman23a.html}
}

@misc{grattafiori2024llama3herdmodels,
      title={The Llama 3 Herd of Models}, 
      author={Aaron Grattafiori and Abhimanyu Dubey and Abhinav Jauhri and Abhinav Pandey and Abhishek Kadian and Ahmad Al-Dahle and Aiesha Letman and Akhil Mathur and Alan Schelten and Alex Vaughan and Amy Yang and Angela Fan and Anirudh Goyal and Anthony Hartshorn and Aobo Yang and Archi Mitra and Archie Sravankumar and Artem Korenev and Arthur Hinsvark and Arun Rao and Aston Zhang and Aurelien Rodriguez and Austen Gregerson and Ava Spataru and Baptiste Roziere and Bethany Biron and Binh Tang and Bobbie Chern and Charlotte Caucheteux and Chaya Nayak and Chloe Bi and Chris Marra and Chris McConnell and Christian Keller and Christophe Touret and Chunyang Wu and Corinne Wong and Cristian Canton Ferrer and Cyrus Nikolaidis and Damien Allonsius and Daniel Song and Danielle Pintz and Danny Livshits and Danny Wyatt and David Esiobu and Dhruv Choudhary and Dhruv Mahajan and Diego Garcia-Olano and Diego Perino and Dieuwke Hupkes and Egor Lakomkin and Ehab AlBadawy and Elina Lobanova and Emily Dinan and Eric Michael Smith and Filip Radenovic and Francisco Guzmán and Frank Zhang and Gabriel Synnaeve and Gabrielle Lee and Georgia Lewis Anderson and Govind Thattai and Graeme Nail and Gregoire Mialon and Guan Pang and Guillem Cucurell and Hailey Nguyen and Hannah Korevaar and Hu Xu and Hugo Touvron and Iliyan Zarov and Imanol Arrieta Ibarra and Isabel Kloumann and Ishan Misra and Ivan Evtimov and Jack Zhang and Jade Copet and Jaewon Lee and Jan Geffert and Jana Vranes and Jason Park and Jay Mahadeokar and Jeet Shah and Jelmer van der Linde and Jennifer Billock and Jenny Hong and Jenya Lee and Jeremy Fu and Jianfeng Chi and Jianyu Huang and Jiawen Liu and Jie Wang and Jiecao Yu and Joanna Bitton and Joe Spisak and Jongsoo Park and Joseph Rocca and Joshua Johnstun and Joshua Saxe and Junteng Jia and Kalyan Vasuden Alwala and Karthik Prasad and Kartikeya Upasani and Kate Plawiak and Ke Li and Kenneth Heafield and Kevin Stone and Khalid El-Arini and Krithika Iyer and Kshitiz Malik and Kuenley Chiu and Kunal Bhalla and Kushal Lakhotia and Lauren Rantala-Yeary and Laurens van der Maaten and Lawrence Chen and Liang Tan and Liz Jenkins and Louis Martin and Lovish Madaan and Lubo Malo and Lukas Blecher and Lukas Landzaat and Luke de Oliveira and Madeline Muzzi and Mahesh Pasupuleti and Mannat Singh and Manohar Paluri and Marcin Kardas and Maria Tsimpoukelli and Mathew Oldham and Mathieu Rita and Maya Pavlova and Melanie Kambadur and Mike Lewis and Min Si and Mitesh Kumar Singh and Mona Hassan and Naman Goyal and Narjes Torabi and Nikolay Bashlykov and Nikolay Bogoychev and Niladri Chatterji and Ning Zhang and Olivier Duchenne and Onur Çelebi and Patrick Alrassy and Pengchuan Zhang and Pengwei Li and Petar Vasic and Peter Weng and Prajjwal Bhargava and Pratik Dubal and Praveen Krishnan and Punit Singh Koura and Puxin Xu and Qing He and Qingxiao Dong and Ragavan Srinivasan and Raj Ganapathy and Ramon Calderer and Ricardo Silveira Cabral and Robert Stojnic and Roberta Raileanu and Rohan Maheswari and Rohit Girdhar and Rohit Patel and Romain Sauvestre and Ronnie Polidoro and Roshan Sumbaly and Ross Taylor and Ruan Silva and Rui Hou and Rui Wang and Saghar Hosseini and Sahana Chennabasappa and Sanjay Singh and Sean Bell and Seohyun Sonia Kim and Sergey Edunov and Shaoliang Nie and Sharan Narang and Sharath Raparthy and Sheng Shen and Shengye Wan and Shruti Bhosale and Shun Zhang and Simon Vandenhende and Soumya Batra and Spencer Whitman and Sten Sootla and Stephane Collot and Suchin Gururangan and Sydney Borodinsky and Tamar Herman and Tara Fowler and Tarek Sheasha and Thomas Georgiou and Thomas Scialom and Tobias Speckbacher and Todor Mihaylov and Tong Xiao and Ujjwal Karn and Vedanuj Goswami and Vibhor Gupta and Vignesh Ramanathan and Viktor Kerkez and Vincent Gonguet and Virginie Do and Vish Vogeti and Vítor Albiero and Vladan Petrovic and Weiwei Chu and Wenhan Xiong and Wenyin Fu and Whitney Meers and Xavier Martinet and Xiaodong Wang and Xiaofang Wang and Xiaoqing Ellen Tan and Xide Xia and Xinfeng Xie and Xuchao Jia and Xuewei Wang and Yaelle Goldschlag and Yashesh Gaur and Yasmine Babaei and Yi Wen and Yiwen Song and Yuchen Zhang and Yue Li and Yuning Mao and Zacharie Delpierre Coudert and Zheng Yan and Zhengxing Chen and Zoe Papakipos and Aaditya Singh and Aayushi Srivastava and Abha Jain and Adam Kelsey and Adam Shajnfeld and Adithya Gangidi and Adolfo Victoria and Ahuva Goldstand and Ajay Menon and Ajay Sharma and Alex Boesenberg and Alexei Baevski and Allie Feinstein and Amanda Kallet and Amit Sangani and Amos Teo and Anam Yunus and Andrei Lupu and Andres Alvarado and Andrew Caples and Andrew Gu and Andrew Ho and Andrew Poulton and Andrew Ryan and Ankit Ramchandani and Annie Dong and Annie Franco and Anuj Goyal and Aparajita Saraf and Arkabandhu Chowdhury and Ashley Gabriel and Ashwin Bharambe and Assaf Eisenman and Azadeh Yazdan and Beau James and Ben Maurer and Benjamin Leonhardi and Bernie Huang and Beth Loyd and Beto De Paola and Bhargavi Paranjape and Bing Liu and Bo Wu and Boyu Ni and Braden Hancock and Bram Wasti and Brandon Spence and Brani Stojkovic and Brian Gamido and Britt Montalvo and Carl Parker and Carly Burton and Catalina Mejia and Ce Liu and Changhan Wang and Changkyu Kim and Chao Zhou and Chester Hu and Ching-Hsiang Chu and Chris Cai and Chris Tindal and Christoph Feichtenhofer and Cynthia Gao and Damon Civin and Dana Beaty and Daniel Kreymer and Daniel Li and David Adkins and David Xu and Davide Testuggine and Delia David and Devi Parikh and Diana Liskovich and Didem Foss and Dingkang Wang and Duc Le and Dustin Holland and Edward Dowling and Eissa Jamil and Elaine Montgomery and Eleonora Presani and Emily Hahn and Emily Wood and Eric-Tuan Le and Erik Brinkman and Esteban Arcaute and Evan Dunbar and Evan Smothers and Fei Sun and Felix Kreuk and Feng Tian and Filippos Kokkinos and Firat Ozgenel and Francesco Caggioni and Frank Kanayet and Frank Seide and Gabriela Medina Florez and Gabriella Schwarz and Gada Badeer and Georgia Swee and Gil Halpern and Grant Herman and Grigory Sizov and Guangyi and Zhang and Guna Lakshminarayanan and Hakan Inan and Hamid Shojanazeri and Han Zou and Hannah Wang and Hanwen Zha and Haroun Habeeb and Harrison Rudolph and Helen Suk and Henry Aspegren and Hunter Goldman and Hongyuan Zhan and Ibrahim Damlaj and Igor Molybog and Igor Tufanov and Ilias Leontiadis and Irina-Elena Veliche and Itai Gat and Jake Weissman and James Geboski and James Kohli and Janice Lam and Japhet Asher and Jean-Baptiste Gaya and Jeff Marcus and Jeff Tang and Jennifer Chan and Jenny Zhen and Jeremy Reizenstein and Jeremy Teboul and Jessica Zhong and Jian Jin and Jingyi Yang and Joe Cummings and Jon Carvill and Jon Shepard and Jonathan McPhie and Jonathan Torres and Josh Ginsburg and Junjie Wang and Kai Wu and Kam Hou U and Karan Saxena and Kartikay Khandelwal and Katayoun Zand and Kathy Matosich and Kaushik Veeraraghavan and Kelly Michelena and Keqian Li and Kiran Jagadeesh and Kun Huang and Kunal Chawla and Kyle Huang and Lailin Chen and Lakshya Garg and Lavender A and Leandro Silva and Lee Bell and Lei Zhang and Liangpeng Guo and Licheng Yu and Liron Moshkovich and Luca Wehrstedt and Madian Khabsa and Manav Avalani and Manish Bhatt and Martynas Mankus and Matan Hasson and Matthew Lennie and Matthias Reso and Maxim Groshev and Maxim Naumov and Maya Lathi and Meghan Keneally and Miao Liu and Michael L. Seltzer and Michal Valko and Michelle Restrepo and Mihir Patel and Mik Vyatskov and Mikayel Samvelyan and Mike Clark and Mike Macey and Mike Wang and Miquel Jubert Hermoso and Mo Metanat and Mohammad Rastegari and Munish Bansal and Nandhini Santhanam and Natascha Parks and Natasha White and Navyata Bawa and Nayan Singhal and Nick Egebo and Nicolas Usunier and Nikhil Mehta and Nikolay Pavlovich Laptev and Ning Dong and Norman Cheng and Oleg Chernoguz and Olivia Hart and Omkar Salpekar and Ozlem Kalinli and Parkin Kent and Parth Parekh and Paul Saab and Pavan Balaji and Pedro Rittner and Philip Bontrager and Pierre Roux and Piotr Dollar and Polina Zvyagina and Prashant Ratanchandani and Pritish Yuvraj and Qian Liang and Rachad Alao and Rachel Rodriguez and Rafi Ayub and Raghotham Murthy and Raghu Nayani and Rahul Mitra and Rangaprabhu Parthasarathy and Raymond Li and Rebekkah Hogan and Robin Battey and Rocky Wang and Russ Howes and Ruty Rinott and Sachin Mehta and Sachin Siby and Sai Jayesh Bondu and Samyak Datta and Sara Chugh and Sara Hunt and Sargun Dhillon and Sasha Sidorov and Satadru Pan and Saurabh Mahajan and Saurabh Verma and Seiji Yamamoto and Sharadh Ramaswamy and Shaun Lindsay and Shaun Lindsay and Sheng Feng and Shenghao Lin and Shengxin Cindy Zha and Shishir Patil and Shiva Shankar and Shuqiang Zhang and Shuqiang Zhang and Sinong Wang and Sneha Agarwal and Soji Sajuyigbe and Soumith Chintala and Stephanie Max and Stephen Chen and Steve Kehoe and Steve Satterfield and Sudarshan Govindaprasad and Sumit Gupta and Summer Deng and Sungmin Cho and Sunny Virk and Suraj Subramanian and Sy Choudhury and Sydney Goldman and Tal Remez and Tamar Glaser and Tamara Best and Thilo Koehler and Thomas Robinson and Tianhe Li and Tianjun Zhang and Tim Matthews and Timothy Chou and Tzook Shaked and Varun Vontimitta and Victoria Ajayi and Victoria Montanez and Vijai Mohan and Vinay Satish Kumar and Vishal Mangla and Vlad Ionescu and Vlad Poenaru and Vlad Tiberiu Mihailescu and Vladimir Ivanov and Wei Li and Wenchen Wang and Wenwen Jiang and Wes Bouaziz and Will Constable and Xiaocheng Tang and Xiaojian Wu and Xiaolan Wang and Xilun Wu and Xinbo Gao and Yaniv Kleinman and Yanjun Chen and Ye Hu and Ye Jia and Ye Qi and Yenda Li and Yilin Zhang and Ying Zhang and Yossi Adi and Youngjin Nam and Yu and Wang and Yu Zhao and Yuchen Hao and Yundi Qian and Yunlu Li and Yuzi He and Zach Rait and Zachary DeVito and Zef Rosnbrick and Zhaoduo Wen and Zhenyu Yang and Zhiwei Zhao and Zhiyu Ma},
      year={2024},
      eprint={2407.21783},
      archivePrefix={arXiv},
      primaryClass={cs.AI},
      url={https://arxiv.org/abs/2407.21783}, 
}

@misc{yang2025qwen3technicalreport,
      title={Qwen3 Technical Report}, 
      author={An Yang and Anfeng Li and Baosong Yang and Beichen Zhang and Binyuan Hui and Bo Zheng and Bowen Yu and Chang Gao and Chengen Huang and Chenxu Lv and Chujie Zheng and Dayiheng Liu and Fan Zhou and Fei Huang and Feng Hu and Hao Ge and Haoran Wei and Huan Lin and Jialong Tang and Jian Yang and Jianhong Tu and Jianwei Zhang and Jianxin Yang and Jiaxi Yang and Jing Zhou and Jingren Zhou and Junyang Lin and Kai Dang and Keqin Bao and Kexin Yang and Le Yu and Lianghao Deng and Mei Li and Mingfeng Xue and Mingze Li and Pei Zhang and Peng Wang and Qin Zhu and Rui Men and Ruize Gao and Shixuan Liu and Shuang Luo and Tianhao Li and Tianyi Tang and Wenbiao Yin and Xingzhang Ren and Xinyu Wang and Xinyu Zhang and Xuancheng Ren and Yang Fan and Yang Su and Yichang Zhang and Yinger Zhang and Yu Wan and Yuqiong Liu and Zekun Wang and Zeyu Cui and Zhenru Zhang and Zhipeng Zhou and Zihan Qiu},
      year={2025},
      eprint={2505.09388},
      archivePrefix={arXiv},
      primaryClass={cs.CL},
      url={https://arxiv.org/abs/2505.09388}, 
}

@inproceedings{
hendrycks2021measuring,
title={Measuring Massive Multitask Language Understanding},
author={Dan Hendrycks and Collin Burns and Steven Basart and Andy Zou and Mantas Mazeika and Dawn Song and Jacob Steinhardt},
booktitle={International Conference on Learning Representations},
year={2021},
url={https://openreview.net/forum?id=d7KBjmI3GmQ}
}

@misc{dolly,
  author = {Databricks},
  title = {Free Dolly: Introducing the World's First Truly Open Instruction-Tuned LLM},
  year = {2023},
  publisher = {GitHub},
  journal = {GitHub repository},
  howpublished = {Blog post},
  url = {https://www.databricks.com/blog/2023/04/12/dolly-first-open-commercially-viable-instruction-tuned-llm}
}

\appendix
\section{Post-hoc Classifier Details}
\label{app:classifier}
Each example is labeled by rules applied to its assistant turn (and its user turn, to detect offered options), in priority order:
(1) a bare yes/no/true/false answer is \textit{binary};
(2) an answer longer than 25 words containing reasoning markers or a trailing ``the answer is (X)'' is \textit{chain of thought}, checked before the letter rule so that rationales ending in a letter are not misrouted;
(3) a bare option letter, with options present in the prompt, is \textit{choice};
(4) a bare number is \textit{numeric};
(5) an answer that parses as JSON is \textit{JSON};
(6) an answer of at most 12 words is \textit{short answer};
(7) anything else is \textit{long generation}.
Numeric is merged into short answer in all reported results, as both are short constrained spans.
On 30 randomly sampled examples labeled by hand, the classifier agreed on 27 (90\%).
Pool base rates are computed from 5{,}000 sampled examples per source, weighted by the true source sizes.

\section{Disattenuation Identity}
\label{app:disatt}
Assume each unit-normalized example gradient of dataset $A$ decomposes as
$g_i = \sqrt{c_A}\,\hat{\mu}_A + \sqrt{1-c_A}\,\eta_i$,
where $\hat{\mu}_A$ is the unit dataset-level direction and the noise terms $\eta_i$ have mean zero and are independent across examples and datasets.
Then for examples $i \neq j$ within $A$, $\mathbb{E}\langle g_i, g_j\rangle = c_A$, so $c_A$ is by construction the expected within-dataset cosine of Equation~\ref{eq:pairmean}.
Across datasets, all noise cross-terms vanish in expectation, leaving
$\mathbb{E}\langle g^A_i, g^B_j\rangle = \sqrt{c_A c_B}\,\cos(\mu_A,\mu_B)$,
which is Equation~\ref{eq:disatt}; dividing the measured mean by $\sqrt{c_A c_B}$ therefore recovers $\cos(\mu_A,\mu_B)$.

\section{Benchmarks and Answer Formats}
\label{app:tables}
Table~\ref{tab:benchmarks} lists the five benchmarks and Table~\ref{tab:format-groups} the seven answer formats and their structure classes.

\begin{table}[h]
\centering
\small
\setlength{\tabcolsep}{4.5pt}
\caption{The five benchmarks we cross with the answer formats of
Table~\ref{tab:format-groups}; each crossing of answer format with a benchmark serves as a \textit{dataset}, totaling 35 datasets.
For the four-option renderings, ARC supplies its own hand-written distractors and GSM8K uses near-miss arithmetic values;
the rest draw distractors from other answers in the same dataset, so the options are always in-distribution.}
\label{tab:benchmarks}
\begin{tabular}{@{}llll@{}}
\toprule
Benchmark & Ability probed & Native answer & Example \\
\midrule
TriviaQA      & open-domain recall      & short entity     & \texttt{PULSAR}             \\
SQuAD         & reading comprehension   & span of passage  & \texttt{1852}               \\
GSM8K         & grade-school math       & integer          & \texttt{109}                \\
ARC-Challenge & science reasoning       & short phrase     & \texttt{a lightning strike} \\
CRUXEval      & code output prediction  & Python value     & \texttt{True}               \\
\bottomrule
\end{tabular}
\end{table}

\begin{table}[h]
\centering
\small
\caption{The seven answer formats we cross with each benchmark, grouped by the structure of the answer the model must produce.}
\label{tab:format-groups}
\begin{tabular}{@{}lll@{}}
\toprule
Structure class & Answer formats & Example answer \\
\midrule
Binary judgment       & True/False, Yes/No       & \texttt{True} \\
Four-option selection & A--D, 1--4, W--Z, a--d   & \texttt{B} \\
Structured generation & JSON                     & \texttt{\{"answer": "Paris"\}} \\
\bottomrule
\end{tabular}
\end{table}

\section{The LESS Selection Pool}
\label{app:pool}
Table~\ref{tab:pool} gives the composition of LESS's 270,679-example selection pool: the four source datasets, their sizes, and the answer-format shares our classifier assigns within each source.
The bottom row is the source-weighted pool base rate, the denominator of every enrichment in Figure~\ref{fig:fig_less_posthoc}.
Note that our classifier labels only 6\% of the CoT source as chain-of-thought format, since rationales without explicit reasoning markers fall under long generation.
Relabeling every CoT-source example as chain-of-thought instead gives BBH 1.7$\times$ on its own format (versus 2.4$\times$ under the rule-based labels), leaves the diagonal of Figure~\ref{fig:fig_less_posthoc} the maximum of every row, and changes no other conclusion.

\begin{table}[h]
\centering
\small
\setlength{\tabcolsep}{4pt}
\caption{Composition of the LESS selection pool \citep{pmlr-v202-longpre23a, NEURIPS2023_949f0f8f, dolly}. Format shares are estimated from 5{,}000 sampled examples per source and weighted by true source sizes. Formats are labeled by the rule-based classifier of Appendix~\ref{app:classifier}; rationales without explicit reasoning markers fall under long generation, so the chain-of-thought base rate is conservative.}
\label{tab:pool}
\begin{tabular}{@{}lrr|rrrrrr@{}}
\toprule
Source & Size & Share & \multicolumn{1}{c}{\shortstack{Multiple\\choice}} & Yes/no & JSON & \multicolumn{1}{c}{\shortstack{Short\\answer}} & \multicolumn{1}{c}{\shortstack{Chain of\\thought}} & \multicolumn{1}{c@{}}{\shortstack{Long\\generation}} \\
\midrule
Flan V2 & 100{,}000 & 36.9\% & 1.6\% & 8.1\% & 0.3\% & 54.4\% & 0.5\% & 35.0\% \\
CoT     & 100{,}000 & 36.9\% & 0.7\% & 0.0\% & 0.0\% & 4.4\%  & 6.4\% & 88.5\% \\
Dolly   & 15{,}011  & 5.5\%  & 0.0\% & 0.2\% & 0.0\% & 25.7\% & 1.4\% & 72.7\% \\
OASST1  & 55{,}668  & 20.6\% & 0.0\% & 0.1\% & 0.0\% & 9.7\%  & 2.6\% & 87.6\% \\
\midrule
Pool (weighted) & 270{,}679 & 100.0\% & 0.9\% & 3.0\% & 0.1\% & 25.1\% & 3.2\% & 67.7\% \\
\bottomrule
\end{tabular}
\end{table}

\begin{figure}[h]
  \centering
  \includegraphics[width=0.85\linewidth]{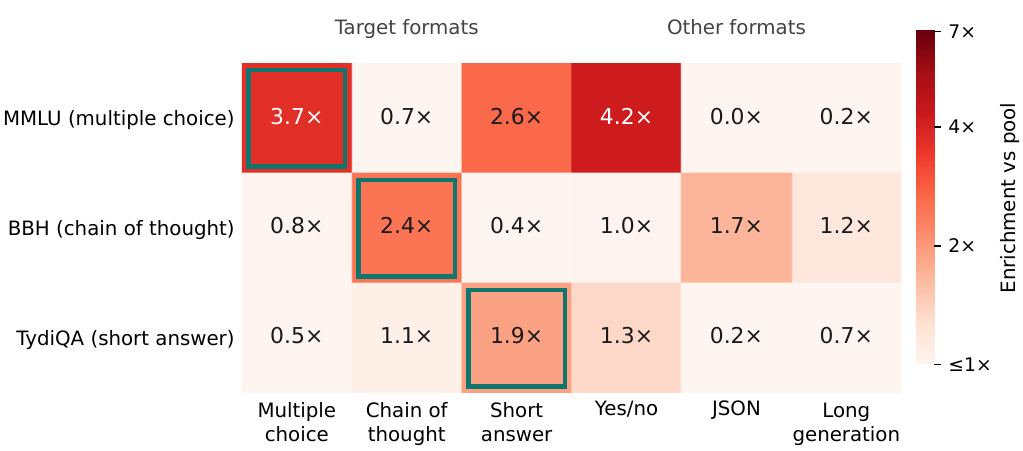}
  \caption{Enrichment of LESS's selections over all six answer formats; the left block is Figure~\ref{fig:fig_less_posthoc}.
  Formats outside each target's structure class are depleted (for MMLU, long generation at 0.2$\times$ and JSON at 0.04$\times$).}
  \label{fig:posthoc_full}
\end{figure}

\section{Aggregation and Error Bars}
\label{app:errorbars}
The trajectory and per-model figures reduce the full grid to two contrasts, and their error bars are designed to answer one question: 
Does the format effect depend on which benchmark was used?
Each contrast is therefore aggregated in two stages, first averaging within a grouping unit defined by benchmark identity, then reporting the mean and $\pm 1$ standard error across those units.

\paragraph{Same class, different task.}
This contrast compares datasets from two different benchmarks that share a format class, so the grouping unit is a benchmark pair.
With five benchmarks there are ten pairs.
Within each pair we average the disattenuated cosine over every same-class format combination (for the pair GSM8K--ARC, for example, GSM8K rendered as A--D against ARC rendered as 1--4, GSM8K as True/False against ARC as Yes/No, and so on), yielding one value per pair; the plotted point is the mean over the ten pairs and the error bar their standard error.

\paragraph{Different class, same task.}
This contrast compares a benchmark with itself rendered in different format classes, so no benchmark pair is involved and the grouping unit is the single benchmark.
With five benchmarks there are five units.
Within each we average over every cross-class format combination (for GSM8K, A--D against True/False, JSON against Yes/No, and so on), and the plotted point is the mean and standard error over the five.

Grouping by benchmark identity is deliberate.
Each cell already averages $64 \times 64$ example pairs, so a standard error over example pairs would be very small and would measure only estimator precision.
The standard error over benchmarks instead measures whether the finding generalizes across task content, which is the uncertainty relevant to our claim; the ten-versus-five asymmetry simply reflects how many benchmark identities each contrast involves.

\section{Compute}
\label{app:compute}
All gradient measurements ran on NVIDIA RTX A6000 GPUs (48\,GB), using at most three concurrently.
Measuring the full grid for one 7B-scale checkpoint takes about one GPU-hour; the 29 reported checkpoints total roughly 30 GPU-hours.
The analysis of LESS's released selections runs on CPU only.

\end{document}